\documentclass[runningheads]{llncs}
\usepackage{tikz}
\usepackage{url,doi}
\usetikzlibrary{calc,arrows.meta}
\usepackage[T1]{fontenc}
\usepackage{graphicx}
\definecolor{darkgreenm}{RGB}{0,100,70}
\definecolor{deepyellow}{RGB}{160,115,20}
\begin{document}
\title{TeRoR: Decoupled Temporal Rotation with Relational Circular Region for Temporal Knowledge Graph Embedding}
%
\titlerunning{TeRoR}
%
\author{Peijia Xie \and Yike Liu \and Chao He \and Huiling Zhu\thanks{Corresponding author.}}
%
%
\institute{South China Normal University, Guangzhou, China}
%
\maketitle              

\begin{abstract}
In recent years, with the emergence of Temporal Knowledge Graphs (TKGs), research on learning entity and relation representations in TKGs has attracted increasing attention, giving rise to a large number of TKG embedding methods. TeRo is a simple and efficient temporal knowledge graph embedding approach. However, TeRo does not do well in modeling the mapping properties of various relations, such as one-to-many, many-to-one, and many-to-many. Meanwhile, it also has limitations in the expression of temporal information. To address these issues, we propose a novel TKG embedding method named TeRoR. This method divides the temporal evolution of entity embeddings, and conducts independent rotation transformations on head and tail entities in the complex vector space to strengthen temporal information modeling capacity. In terms of relational characteristics, we train a radius $\rho_r$ to constrain the rotated and translated head entities within a circular region centered on the tail entity, which effectively captures the diverse mapping properties of relations. Experimental results demonstrate that TeRoR achieves competitive performance against state-of-the-art models on four distinct TKG datasets.

\keywords{Temporal knowledge graphs \and Rotation \and Complex vector space \and Link prediction.}
\end{abstract}
\section{Introduction}
In recent years, knowledge graphs (KGs) have emerged as large-scale structured semantic repositories and become indispensable infrastructure for diverse artificial intelligence applications, ranging from question answering and information retrieval to recommendation systems and cognitive reasoning. Classic publicly accessible KGs, including DBpedia~\cite{auer2007dbpedia}, YAGO~\cite{fabian2007yago}, NELL~\cite{carlson2010toward}, and Freebase~\cite{bollacker2008freebase}, store real-world entities and their interconnections in the form of directed graphs. In general, factual knowledge in KGs is standardized as triple notation $(s,r,o)$, where the subject $s$ and object $o$ correspond to graph nodes (entities), and the relation $r$ acts as the edge connecting paired entities. Nevertheless, real-world KGs typically feature massive data volume and heterogeneous attribute distribution, accompanied by widespread data incompleteness and missing relational facts. To tackle such defects and improve KG completeness, knowledge graph embedding (KGE) techniques have been rapidly developed. These technologies project discrete symbolic entities and relations into continuous low-dimensional vector spaces, and adopt customized scoring functions to measure the rationality of triples. Among existing KGE algorithms, translation-based models represented by TransE~\cite{bordes2013translating}, TransH~\cite{wang2014knowledge}, TransR~\cite{lin2015learning}, and TransD~\cite{ji2015knowledge} have gained widespread attention due to their balanced performance in model complexity and prediction accuracy.

Although static KGE models have achieved promising results in link prediction tasks, their core assumption that all relational facts hold permanently and universally contradicts real-world scenarios. In fact, most factual relations in practical KGs are time-sensitive and dynamically evolving. For example, the relational fact (Obama, President of, USA) is only valid from 2009 to 2017, while the event (Einstein, Died in, Princeton) uniquely occurred in 1955. To record such time-dependent facts, temporal knowledge graphs (TKGs) including ICEWS~\cite{lautenschlager2015icews}, GDELT~\cite{leetaru2013gdelt}, and YAGO3~\cite{mahdisoltani2014yago3} have been constructed, which extend the traditional triple structure into quadruples with additional timestamps or time intervals. The prevalence of TKGs further promotes the research of temporal knowledge graph embedding (TKGE). However, directly applying static KGE models to TKGs usually leads to severe performance degradation, since these static methods ignore temporal characteristics, especially when modeling transient relations such as visits, tenure, and commodity transactions.

To address the above limitation, scholars have proposed various TKGE approaches to embed temporal information into entity and relation representations through explicit or implicit encoding strategies. Typical methods such as TTransE~\cite{leblay2018deriving} and TA-DistMult~\cite{garcia2018learning} have proven superior to conventional static KGE models in temporal link prediction tasks. Even so, current TKGE frameworks still have prominent drawbacks. First, most existing TKGE models are simple temporal extensions based on TransE or DistMult, lacking sufficient expressive capability to capture intricate temporal dependencies. Second, several temporal embedding methods adopt time-specific hyperplane structures to model temporal variation, while inherently assuming that most temporal relations are one-to-one mappings. Such an assumption cannot adapt to the diverse complex relations (many-to-one, one-to-many, and many-to-many) commonly existing in real-world TKGs. Accordingly, designing a high-performance TKGE framework that can accurately capture temporal evolution and flexibly fit multi-type relational mappings remains a challenging research problem.

To overcome these challenges, this paper proposes a novel model named TeRoR based on the model TeRo~\cite{xu2020tero}. The innovations of TeRoR lie in two core designs. On the one hand, the model decouples the temporal influence of timestamps on subject and object entities in quadruples and performs independent temporal evolution, which strengthens the utilization and expression ability of temporal information. On the other hand, relation information is used to parameterize circular validity regions for positive quadruples, enabling the model to precisely fit different types of multi-relational interactions. Sufficient experimental results on mainstream benchmarks verify that the proposed TeRoR outperforms state-of-the-art baseline models in temporal link prediction. The major contributions of this paper are summarized as follows:

A novel TKGE architecture with decoupled temporal evolution for subject and object entities is proposed, which greatly enhances the temporal information representation capability of embedding models.

A relation-aware circular region is designed as the plausibility boundary of valid quadruples, which realizes effective modeling of complex multi-relational interactions in TKGs.

Systematic experiments are conducted on multiple public TKG datasets. The empirical results demonstrate that TeRoR achieves stable and significant performance improvements over advanced baseline models in key evaluation metrics, including MRR and Hits@K.

\section{Related Work}
This section comprehensively reviews representative static knowledge graph embedding (KGE) methods and state-of-the-art temporal knowledge graph embedding (TKGE) techniques, with an in-depth discussion on the inherent limitations of existing embedding frameworks.
 
Conventional static KGE models are generally divided into two mainstream categories: translation-based distance measurement models and semantic matching models. 

Distance-based models quantify the credibility of relational triples by calculating geometric vector distances after relational transformation. As a seminal translation-based method, TransE~\cite{bordes2013translating} treats relations as translational vectors within a unified embedding space. Nevertheless, it exhibits weak modeling capability for complex multi-relational patterns, including one-to-many, many-to-one and many-to-many interactions. To mitigate this defect, improved variants such as TransH~\cite{wang2014knowledge}, TransR~\cite{lin2015learning}, and TransD~\cite{ji2015knowledge} have been proposed. These models project entities into relation-dependent subspaces and adopt customized mapping matrices to enhance representation flexibility. Furthermore, RotatE~\cite{sun2019rotate} implements rotational transformation in complex space to simulate relational properties.

In contrast, semantic matching models evaluate factual rationality by measuring latent semantic compatibility between entities and relations. Typical implementations cover tensor factorization-based RESCAL~\cite{nickel2011three} and holographic correlation-based HoLE~\cite{nickel2016holographic}. DistMult~\cite{yang2014embedding} simplifies calculation via diagonal relational matrices, yet its symmetric structural design makes it unable to capture asymmetric relations. To address this issue, ComplEx~\cite{trouillon2016complex} and QuatE~\cite{zhang2019quaternion} extend embeddings to complex and quaternion spaces, enabling effective modeling of asymmetric and diverse static relational patterns. However, all the above models follow a static learning paradigm without dedicated temporal modeling modules, which severely restricts their applicability for dynamic temporal knowledge graphs.

To capture time-varying factual characteristics, TKGE algorithms embed temporal cues into entity and relation representation learning. Current TKGE techniques can be classified into two primary technical branches:

Translation-based temporal extensions. This category evolves from the classic TransE architecture. Representative methods including TTransE~\cite{leblay2018deriving}, TA-TransE~\cite{garcia2018learning}, HyTE~\cite{dasgupta2018hyte}, and ATiSE~\cite{xu2020temporal} realize temporal modeling through time embedding addition, attention allocation, and time-aware hyperplane projection. Specifically, HyTE~\cite{dasgupta2018hyte} builds independent temporal hyperplanes and maps entities to time slices to reflect dynamic temporal characteristics.

Semantic matching-based temporal extensions. These approaches are mainly improved on the basis of DistMult~\cite{yang2014embedding}. For example, TA-DistMult~\cite{garcia2018learning} incorporates temporal information into the matching process. DE-SimplE~\cite{goel2020diachronic} generates diachronic entity embeddings to capture time-evolving features, which achieves favorable performance on diverse relation types.

Despite the progress achieved by existing TKGE methods, two critical drawbacks still remain. Firstly, most models adopt simplistic temporal encoding strategies and fail to capture fine-grained temporal dependencies. Secondly, few algorithms can effectively fit complex multi-relational mappings (1-N, N-1, N-N) in dynamic temporal scenarios. Such unresolved bottlenecks motivate us to construct a robust temporal embedding model with stronger temporal perception and relational modeling capability.

\section{Method}
This section presents the detailed methodology of TeRoR, a novel knowledge graph embedding model grounded in the temporal rotation mechanism within the complex vector space. As an improved iteration of the classical temporal embedding model TeRo, TeRoR is developed to address two critical inherent drawbacks of the baseline model. To be specific, TeRo suffers from insufficient temporal feature utilization and weak modeling performance for sophisticated multi-relational patterns, such as one-to-many, many-to-one, and many-to-many interactions. For logical clarity, we first revisit the fundamental principles and latent deficiencies of TeRo, before elaborating on the detailed architectural design of TeRoR.
\subsection{TeRo~\cite{xu2020tero}}
TeRo implements temporal evolution modeling by rotating entity embeddings in the complex domain, which facilitates the embedding representation and logical inference of time-dependent factual quadruples. In this work, $\mathcal{E}$, $\mathcal{R}$, and $\mathcal{T}$ are defined as the entity set, relation set, and timestamp set, respectively. A temporal knowledge graph is composed of numerous quadruples formulated as $(s,r,o,t)$, where $s, o \in \mathcal{E}$ correspond to the subject and object entities, $r \in \mathcal{R}$ denotes the relational semantics, and $t$ is the timestamp marking the occurrence of a certain fact. Each valid quadruple $(s,r,o,t)$ is assigned an individual time step $\tau \in \mathcal{T}$. In TeRo, time-evolving entity embeddings are generated by applying temporal rotation operations to static entity vectors. The corresponding transformation function is formulated as follows:
\begin{equation}
	s_t = s \circ \tau, \quad o_t = o \circ \tau 
\end{equation}
where $\circ$ refers to the Hermitian dot product between two complex vectors. Nevertheless, such a temporal rotation paradigm can only capture limited temporal features, resulting in incomplete temporal information perception.

For any positive quadruple $(s,r,o,t) \in Q^+$, TeRo treats the relation embedding $r$ as a translational vector connecting the time-aware subject embedding $s_t$ and the conjugate of time-aware object embedding $\bar{o_t}$. Here, $r \in \mathcal{R}$, and $Q^+$ represents the collection of all valid positive quadruples. The scoring function adopted in TeRo is defined as:
\begin{equation}
	score(s,r,o,t)_{TeRo} =  \left\| s_t + r - \bar{o_t} \right\|
\end{equation}

Despite its effective temporal rotation design, the above scoring function still has obvious limitations. It cannot precisely fit and distinguish diverse complex relational mappings, which restricts its modeling capacity for one-to-many, many-to-one, and many-to-many relational scenarios prevalent in real-world TKGs.
\subsection{TeRoR}
To adress the aforementioned two defects of TeRo, we propose an improved temporal embedding model TeRoR.

\subsubsection{Representation of temporal information}Different from the shared single time step adopted in TeRo, TeRoR decouples temporal representation into two independent phase rotations, which are separately assigned to subject $s$ and object $o$ to strengthen temporal evolution. For an arbitrary timestamp $t$, two exclusive time variables $\tau_s$ and $\tau_o$ are utilized to implement asynchronous rotation for subject and object entities. The temporal evolution formula is redefined as:
\begin{equation}
	s_t = s \circ \tau_s, \quad o_t = o \circ \tau_o 
\end{equation}
where $\circ$ represents the Hermitian dot product for complex vectors. In accordance with Euler's formula $e^{i\theta} = \cos\theta + i\sin\theta$, unit complex numbers can be geometrically interpreted as rotational transformations on the complex plane. To satisfy this geometric constraint, every element within the time-related vectors $\tau_s, \tau_o \in \mathcal{C}^d$ is restricted with a unit modulus, namely $|\tau_{s,j}| = 1$ and $|\tau_{o,j}| = 1$. Mathematically, each element $\tau_{s,j}$ can be expressed as $e^{i\theta_{s,j}}$, which corresponds to a counterclockwise rotation with an angle of $\theta_{s,j}$ for the subject entity $s$. The object entity $o$ follows the identical rotation rule governed by $\tau_o$. This rotation operation only modifies the phase of entity embeddings while keeping their modulus unchanged.

\subsubsection{Relational mapping properties}To effectively characterize complex relational mappings (i.e., one-to-many, many-to-one, and many-to-many), a learnable radius vector $\rho_r \in \mathcal{C}^d$ is introduced for each relation. This radius parameter constructs a circular valid region to constrain the reasonable distribution of positive quadruples. The scoring calculation is formulated as:
\begin{equation}
	dis =  \left\| s_t + r - \bar{o_t} \right\|
\end{equation}
\begin{equation}
	l(s,r,o,t) =  max(dis, k_t \circ \rho_r)
\end{equation}
where $k_t \in \mathcal{C}^d$ denotes a time-dependent scaling factor that dynamically adjusts the valid radius $\rho_r$ at different timestamps.
\begin{figure}[h]
	\centering
	\begin{tikzpicture}[>=Stealth, scale=1.2]
		\draw[->, gray] (-2.5,0) -- (2.5,0);
		\draw[->, gray] (0,-3) -- (0,2.5);
		
		\coordinate (O) at (0,0);
		\coordinate (o) at (-1.5,0);
		\coordinate (o_t) at (-0.45,1.5);
		\coordinate (s) at (0,2);
		\coordinate (s_t) at ({sqrt(3)},1);
		\coordinate (o_t_bar) at (-0.45,-1.5);
		\coordinate (s_t^{'}) at (0.1,-1);
		
		\foreach \p in {o, o_t, o_t_bar}
		\fill[blue!60] (\p) circle (1.5pt);
		\foreach \s in {s, s_t, s_t^{'}}
		\fill[darkgreenm!60] (\s) circle (1.5pt);
		\draw[orange, dashed, thick] (o_t_bar) circle (1);
		\draw[thick] (o_t_bar) -- ($(o_t_bar) + (1,0)$) node[midway, below=1pt] {$k_t \circ \rho_r$};

		\draw[thick,->,blue] (O) -- (o_t) node[midway, above left] {$o_{t}$};
		\draw[thick,->,blue] (O) -- (o);
		\draw[thick,->,darkgreenm] (O) -- (s_t) node[midway, above=2pt] {$s_{t}$};
		\draw[thick,->,darkgreenm] (O) -- (s);
		\draw[thick,->,blue] (O) -- (o_t_bar) node[midway, below left] {$\overline{o}_{t}$};
		\draw[thick,->,deepyellow] (s_t) -- (s_t^{'}) node[midway, below right] {$r$};
		\draw[dashed] (o_t_bar) -- (s_t^{'}) node[midway, below right=-3pt] {$dis$};
		
		\draw[dashed, gray, ->] (o) to[bend left=30] node[midway, above=3pt] {$\tau_o$} (o_t);
		\draw[dashed, gray, ->] (s) to[bend left=30] node[midway, above=3pt] {$\tau_s$} (s_t);
		
		\node at (s) [left,darkgreenm] {$s$};
		\node at (o) [below,blue] {$o$};
		\node at (s_t^{'}) [right,darkgreenm] {$s_t^{'}$};
	\end{tikzpicture}
	\caption{Illustration of TeRoR with only one embedding dimension.}
\end{figure}
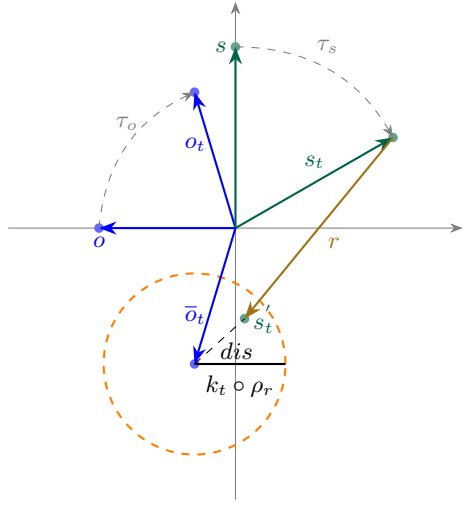

For time-interval facts in TKGs, the temporal annotation is defined as $t = [t_b, t_e]$, where $t_b$ and $t_e$ separately represent the start and end moments of a continuous fact. Following the decomposition paradigm of TeRo, such time-interval quadruples are split into two independent time-point samples: $(s, r_b, o, t_b)$ and $(s, r_e, o, t_e)$. Accordingly, the original relation set is extended into two disjoint sets $R_b$ and $R_e$, where $r_b \in R_b$ denotes the initial relational state and $r_e \in R_e$ represents the termination state. The final score of the time-interval fact is computed as the average value of the two decomposed quadruples to comprehensively reflect the temporal boundary characteristics:
\begin{equation}
	l(s,r,o,[t_b, t_e]) =  \frac{1}{2}(l(s_{t_b},r_b,o_{t_b},t) + l(s_{t_e},r_e,o_{t_e},t))
\end{equation}

Moreover, the proposed TeRoR is adaptable to incomplete temporal annotation scenarios. For facts with only a start timestamp $(s, r, o, [t_b, -])$ or an end timestamp $(s, r, o, [-, t_e])$, their scores are determined by the single available time-aware quadruple. The simplified calculation rules are given by
\begin{equation}
	l(s, r, o, [t_b, -])=l(s, r_b, o, t_b)
\end{equation}
\begin{equation}
	l(s, r, o, [-, t_e])=l(s, r_e, o, t_e)
\end{equation}

\subsubsection{In terms of model optimization}
This work adopts the negative sampling loss function proposed in~\cite{sun2019rotate}, which maintains consistent optimization settings with the original TeRo~\cite{xu2020tero} model. This loss function has been proven effective for distance-based embedding models and has been widely applied in classic KGE models including RotatE~\cite{sun2019rotate} and ATiSE~\cite{xu2020temporal}. The loss function is defined as:
\begin{equation}
	Loss(\xi) = -\log \sigma(\gamma - l(\xi)) - \sum_{i=1}^{\eta} \frac{1}{\eta} \log \sigma(l(\xi_i') - \gamma)
\end{equation}
where $\xi \in Q^+$ represents a positive quadruple, and $\xi_i'$ denotes the $i$-th negative sample generated by randomly corrupting the subject or object entity. $\sigma(\cdot)$ is the Sigmoid activation function, $\gamma$ refers to the fixed margin hyperparameter, and $\eta$ controls the quantity of negative samples during training.
\section{Experiments}
\subsection{Temporal Knowledge Graph Datasets}
We selected four datasets: ICEWS14, ICEWS05-15, YAGO11k, and Wikidata12k. These datasets are widely used benchmarks in Temporal Knowledge Graph (TKG) embedding research, originating from diverse knowledge sources and covering both event-based facts and general encyclopedic knowledge.

ICEWS14 and ICEWS05-15 are both derived from the Integrated Conflict Early Warning System (ICEWS), which records global political and military events. ICEWS14 contains events from the entire year of 2014 and is relatively small in scale, making it suitable for rapid model validation. ICEWS05-15 spans eleven years from 2005 to 2015, featuring longer temporal coverage and a larger number of entities and events, thereby enabling a more rigorous assessment of models' ability to capture long-term temporal dependencies. Both datasets represent events as quadruples $(s, p, o, t)$ with a daily timestamp granularity.

YAGO11k is a subset extracted from the YAGO knowledge base, primarily comprising temporal facts related to historical figures and locations, such as birth/death dates and the start/end times of positions. Its temporal information is mostly expressed as time intervals or discrete years. With approximately 11,000 entities and encyclopedic-style relations, this dataset is well-suited for evaluating models' capacity to represent time intervals.

Wikidata12k is sourced from Wikidata and filters 12,000 entities along with their associated facts that possess explicit timestamps. Similar to YAGO11k, it contains a substantial amount of interval-based temporal data; however, it covers broader domains, including geography, organizations, and creative works. This dataset is commonly employed to verify models' generalization performance on sparse temporal knowledge.

In summary, the ICEWS series leans toward event prediction scenarios with fine-grained timestamps and high fact density, whereas YAGO11k and Wikidata12k are oriented toward encyclopedic fact completion, featuring coarse-grained timestamps and a predominance of interval-based data.Table 1 summarizes the statistics of the four datasets used in our model.
\begin{table}[h]
	\caption{Statistics of datasets.}
	\centering
	\begin{tabular}{|l|c|c|c|c|c|c|}
		\hline
		Dataset      & Entities & Relations & Time Span  & Training & Validation & Test   \\
		\hline
		ICEWS14      & 6,869      & 230         & 2014       & 72,826     & 8,941       & 8,963    \\
		ICEWS05-15   & 10,094     & 251         & 2005-2015  & 368,962    & 46,275      & 46,092   \\
		YAGO11k      & 10,623     & 10          & -453-2844  & 16,406     & 2,050       & 2,051    \\
		Wikidata12k  & 12,554     & 24          & 1479-2018  & 32,497     & 4,062       & 4,062    \\
		\hline
	\end{tabular}
	\label{tab:datasets}
\end{table}

\subsection{Evaluation Metrics}
Two commonly adopted evaluation metrics are MRR (Mean Reciprocal Ranking) and Hits@K. The core of MRR is to calculate the average of the reciprocal ranks of correct triples among all candidate results. If the correct entity ranks first, its reciprocal rank is 1; if it ranks third, the reciprocal rank is 1/3. The higher the ranking, the closer the MRR value is to 1. It comprehensively reflects the overall ranking capability of the model, is sensitive to incorrectly ranked samples with low rankings, and can objectively demonstrate the global reasoning performance of the model.Hits@K denotes the proportion of samples where the correct entity appears within the top-K positions of the ranking list, with Hit@1, Hit@3 and Hit@10 being the most frequently used. Hit@1 represents the accuracy of exact matching, while Hit@10 evaluates whether the model can place the correct result in the top candidate positions. A higher value indicates better retrieval and recall performance of the model.

\subsection{Baselines}
To comprehensively validate the effectiveness of our model, we selected two categories of baselines: static KGE methods, including TransE~\cite{bordes2013translating}, DistMult~\cite{yang2014embedding}, ComplEx-N3~\cite{lacroix2018canonical}, RotatE~\cite{sun2019rotate} and QuatE~\cite{zhang2019quaternion}; and temporal KGE methods, including TTransE~\cite{leblay2018deriving}, TA-TransE~\cite{garcia2018learning}, TA-DistMult~\cite{garcia2018learning}, DE-SimplE~\cite{goel2020diachronic}, ATiSE~\cite{xu2020temporal} and TeRo~\cite{xu2020tero}. The baseline results are primarily adopted from the studies of~\cite{goel2020diachronic} and~\cite{xu2020tero}, which employed the same evaluation protocol as ours.

\subsection{Experimental Setup}
The experimental setup follows the settings of TeRo, using MRR on the validation set as the early stopping criterion to select the optimal hyperparameters. The maximum number of training epochs is set to 5000, and the batch size is uniformly fixed at 512 for all datasets. The embedding dimension $d$, the number of negative samples $\eta$, the margin $\gamma$, and the learning rate r are determined via grid search over the candidate sets {100, 200, ..., 1000}, {1, 3, 5, 10}, {1, 2, 3, 5, ..., 120}, and {1, 0.3, 0.1, 0.03, 0.01}, respectively. We adopt Adagrad as the optimizer. The final non-default parameters for each dataset are as follows: ICEWS14 uses $d = 500$,$lr = 0.1$ and $\gamma = 50$; ICEWS05-15 uses $d = 500$, $lr = 0.1$ and $\gamma = 50$; YAGO11k uses $d = 500$, $lr = 0.1$ and $\gamma = 50$; Wikidata12k uses $d = 700$, $lr = 0.1$ and $\gamma = 10$.
\section{Results and Analysis}
\subsection{Comparative study}
\begin{table}[htbp]
	\caption{Results on ICEWS14 and ICEWS05-15. *: results are taken from~\cite{garcia2018learning}. $\diamond$: results are taken from~\cite{goel2020diachronic}. $\dagger$: results are taken from~\cite{xu2020tero}.
		Dashes: results are unobtainable. The best results among all models are written bold. The second-best results are underlined.}	\label{tab:result_icews}
	\centering
	\begin{tabular}{|c|c|c|c|c|c|c|c|c|}
		\hline
		Datasets & \multicolumn{4}{c|}{ICEWS14} & \multicolumn{4}{c|}{ICEWS05-15} \\
		\hline
		Metrics & MRR & Hits@1 & Hits@3 & Hits@10 & MRR & Hits@1 & Hits@3 & Hits@10 \\
		\hline
		TransE$^*$ & .280 & .094 & - & .637 & .294 & .090 & - & .663 \\
		DistMult$^*$ & .439 & .323 & - & .672 & .456 & .337 & - & .691 \\
		ComplEx-N3$^\dagger$ & .467 & .347 & .527 & .716 & .481 & .362 & .535 & .729 \\
		RotatE$^\dagger$ & .418 & .291 & .478 & .690 & .304 & .164 & .355 & .595 \\
		QuatE$^2$ $^\dagger$ & .471 & .353 & .530 & .712 & .482 & .370 & .529 & .727 \\
		\hline
		TTransE$^\diamond$ & .255 & .074 & - & .601 & .271 & .084 & - & .616 \\
		HyTE$^\diamond$ & .297 & .108 & .416 & .655 & .316 & .116 & .445 & .681 \\
		TA-TransE$^*$ & .275 & .095 & - & .625 & .299 & .096 & - & .668 \\
		TA-DistMult$^*$ & .477 & .363 & - & .686 & .474 & .346 & - & .728 \\
		DE-SimplE$^\diamond$ & .526 & .418 & .592 & .725 & .513 & .392 & .578 & .748 \\
		ATiSE$^\dagger$ & .550 & .436 & \textbf{.629} & \textbf{.750} & .519 & .378 & .606 & .794 \\
		TeRo$^\dagger$ & \underline{.562} & \underline{.468} & .621 & .732 & \underline{.586} & \underline{.469} & \underline{.668} & \underline{.795} \\
		\hline
		Ours & \textbf{.572} & \textbf{.482} & \underline{.626} & \underline{.741} & \textbf{.612} & \textbf{.505} & \textbf{.687} & \textbf{.797} \\
		\hline
	\end{tabular}
\end{table}
\begin{table}[htbp]
	\centering
	\caption{Results on YAGO11k and Wikidata12k. $\dagger$: results are taken from~\cite{xu2020tero}. The best results among all models are written bold. The second-best results are underlined.}
	\label{tab:result_yago_wiki}
	\begin{tabular}{|c|c|c|c|c|c|c|c|c|}
		\hline
		Datasets & \multicolumn{4}{c|}{YAGO11k} & \multicolumn{4}{c|}{Wikidata12k} \\
		\hline
		Metrics & MRR & Hits@1 & Hits@3 & Hits@10 & MRR & Hits@1 & Hits@3 & Hits@10 \\
		\hline
		TransE$^\dagger$ & .100 & .015 & .138 & .244 & .178 & .100 & .192 & .339 \\
		DistMult$^\dagger$ & .158 & .107 & .161 & .268 & .222 & .119 & .238 & .460 \\
		ComplEx-N3$^\dagger$ & .167 & .106 & .154 & .282 & .233 & .123 & .253 & .436 \\
		RotatE$^\dagger$ & .167 & .103 & .167 & .305 & .221 & .116 & .236 & .461 \\
		QuatE$^2$ $^\dagger$ & .164 & .107 & .148 & .270 & .230 & .125 & .243 & .416 \\
		\hline
		TTransE$^\dagger$ & .108 & .020 & .150 & .251 & .172 & .096 & .184 & .329 \\
		HyTE$^\dagger$ & .105 & .015 & .143 & .272 & .180 & .098 & .197 & .333 \\
		TA-TransE$^\dagger$ & .127 & .027 & .160 & .326 & .178 & .030 & .267 & .429 \\
		TA-DistMult$^\dagger$ & .161 & .103 & .171 & .292 & .218 & .122 & .232 & .447 \\
		ATiSE$^\dagger$ & .170 & .110 & .171 & .288 & .280 & .175 & .317 & .481 \\
		TeRo$^\dagger$ & \underline{.187} & \textbf{.121} & \textbf{.197} & \underline{.319} & \textbf{.299} & \textbf{.198} & \textbf{.329} & \textbf{.507} \\
		\hline
		Ours & \textbf{.189} & \textbf{.121} & \textbf{.197} & \textbf{.326} & \underline{.288} & \underline{.190} & \underline{.320} & \underline{.487} \\
		\hline
	\end{tabular}
\end{table}
Table 1 and Table 2 report the link prediction performance of TeRoR and baseline models on four typical TKG datasets. As observed in the tables, temporal embedding models generally outperform static KGE methods owing to their ability to capture time-evolving factual information. Owning to the decoupled temporal rotation and relation-defined circular valid region, TeRoR achieves superior overall performance. Specifically, compared with TeRo, TeRoR increases the MRR by 1.0\% (0.562→0.572) on ICEWS14 and 2.6\% (0.586→0.612) on ICEWS05-15, with consistent gains in Hits@K metrics. TeRoR achieves the best MRR of 0.189 on YAGO11k. The slight performance lag on Wikidata12k stems from data sparsity, which restricts the optimization of radius parameters. Overall, TeRoR reliably captures temporal evolution and complex relations, exhibiting prominent advantages on dense TKGs.

\subsection{Ablation study}
\begin{table}[htbp]
	\centering
	\caption{Ablation study results on ICEWS14 and ICEWS05-15. (A): independent entity rotation. (B): radius-constrained circular region. The best results are written bold.}
	\label{tab:ablation_icews}
	\setlength{\tabcolsep}{4pt}
	\begin{tabular}{|c|c|c|c|c|c|c|c|c|}
		\hline
		Datasets & \multicolumn{4}{c|}{ICEWS14} & \multicolumn{4}{c|}{ICEWS05-15} \\
		\hline
		Metrics & MRR & Hits@1 & Hits@3 & Hits@10 & MRR & Hits@1 & Hits@3 & Hits@10 \\
		\hline
		TeRo & .562 & .468 & .621 & .732 & .586 & .469 & .668 & .795 \\
		Ours(A) & .572 & .481 & \textbf{.630} & .736 & .608 & .498 & .687 & \textbf{.800} \\
		Ours(A+B) & \textbf{.572} & \textbf{.482} & .626 & \textbf{.741} & \textbf{.612} & \textbf{.505} & \textbf{.687} & .797 \\
		\hline
	\end{tabular}
\end{table}

\begin{table}[htbp]
	\centering
	\caption{Ablation study results on YAGO11k and Wikidata12k. (A): independent entity rotation. (B): radius-constrained circular region. The best results are written bold.}
	\label{tab:ablation_yago_wiki}
	\setlength{\tabcolsep}{4pt}
	\begin{tabular}{|c|c|c|c|c|c|c|c|c|}
		\hline
		Datasets & \multicolumn{4}{c|}{YAGO11k} & \multicolumn{4}{c|}{Wikidata12k} \\
		\hline
		Metrics & MRR & Hits@1 & Hits@3 & Hits@10 & MRR & Hits@1 & Hits@3 & Hits@10 \\
		\hline
		TeRo & .178 & .121 & .197 & .319 & \textbf{.299} & \textbf{.198} & \textbf{.329} & .507 \\
		Ours(A) & .186 & .116 & .195 & .329 & .296 & .193 & \textbf{.329} & \textbf{.509} \\
		Ours(A+B) & \textbf{.189} & \textbf{.121} & \textbf{.197} & \textbf{.326} & .288 & .190 & .320 & .487 \\
		\hline
	\end{tabular}
\end{table}
Ablation studies on the ICEWS14, ICEWS05-15, YAGO11k and Wikidata12k datasets reveal that the performance gains of TeRoR primarily stem from two key modules: independent entity rotation (A) and the radius-constrained circular region (B).

First, when only module A is incorporated (TeRoR (A)), the model already outperforms the baseline TeRo across all metrics. On ICEWS14, MRR improves from 0.562 to 0.572 and Hits@1 rises from 0.468 to 0.481; on ICEWS05-15, MRR increases from 0.586 to 0.608 and Hits@10 grows from 0.795 to 0.800. Performance improvements are also observed on YAGO11k and Wikidata12k. This indicates that performing independent phase rotations for $s$ and $o$ in the complex plane enables more flexible modeling of temporal dependencies and significantly enhances the ranking quality of link prediction.

Upon further integrating module B (TeRoR (A+B)), model performance is boosted even further. On ICEWS05-15, MRR reaches 0.612 and Hits@1 climbs to 0.505; on YAGO11k, MRR improves to 0.189 and Hits@10 increases to 0.326. This demonstrates that the radius-constrained circular domain provides more precise geometric boundaries for entity embeddings, effectively strengthening the model's capacity to model complex relations and temporal information.

In summary, both modules A and B contribute significantly to model performance. Their combination maximizes the advantages of complex space rotational embedding, thereby validating the effectiveness of each component in the TeRoR design.

\subsection{Relation mapping properties}
\begin{table}[htbp]
	\caption{Performance on ICEWS14 by relation mapping properties. The best results are written bold.} \label{tab:type_icews14}
	\centering
	\resizebox{\linewidth}{!}{%
		\begin{tabular}{|l|c|c|c|c|c|c|c|c|c|c|c|c|c|c|c|c|}
			\hline
			& \multicolumn{4}{c|}{1-1} & \multicolumn{4}{c|}{1-N} & \multicolumn{4}{c|}{N-1} & \multicolumn{4}{c|}{N-N} \\
			\hline
			& MRR & H@1 & H@3 & H@10 & MRR & H@1 & H@3 & H@10 & MRR & H@1 & H@3 & H@10 & MRR & H@1 & H@3 & H@10 \\
			\hline
			TeRo   & .498 & .420 & .545 & .640 & .630 & .539 & .691 & .796 & .376 & .284 & \textbf{.424} & \textbf{.557} & .630 & .520 & .703 & .826 \\
			Ours  & \textbf{.506} & \textbf{.431} & \textbf{.551} & \textbf{.651} & \textbf{.657} & \textbf{.572} & \textbf{.709} & \textbf{.817} & \textbf{.385} & \textbf{.300} & .415 & .552 & \textbf{.653} & \textbf{.549} & \textbf{.719} & \textbf{.843} \\
			\hline
		\end{tabular}
	}
\end{table}

\begin{table}[htbp]
	\centering
	\caption{Performance on ICEWS05-15 by relation mapping properties. The best results are written bold.}
	\label{tab:type_icews05}
	\resizebox{\linewidth}{!}{%
		\begin{tabular}{|l|c|c|c|c|c|c|c|c|c|c|c|c|c|c|c|c|}
			\hline
			& \multicolumn{4}{c|}{1-1} & \multicolumn{4}{c|}{1-N} & \multicolumn{4}{c|}{N-1} & \multicolumn{4}{c|}{N-N} \\
			\hline
			& MRR & H@1 & H@3 & H@10 & MRR & H@1 & H@3 & H@10 & MRR & H@1 & H@3 & H@10 & MRR & H@1 & H@3 & H@10 \\
			\hline
			TeRo   & .475 & .398 & .521 & .616 & .658 & .575 & .713 & .799 & .358 & .251 & .415 & .562 & .614 & .478 & .713 & .851 \\
			Ours  & \textbf{.491} & \textbf{.419} & \textbf{.530} & \textbf{.625} & \textbf{.670} & \textbf{.596} & \textbf{.719} & \textbf{.803} & \textbf{.395} & \textbf{.296} & \textbf{.447} & \textbf{.580} & \textbf{.640} & \textbf{.517} & \textbf{.731} & \textbf{.852} \\
			\hline
		\end{tabular}
	}
\end{table}

\begin{table}[htbp]
	\centering
	\caption{Performance on YAGO11k by relation mapping properties. The best results are written bold.}
	\label{tab:type_yago}
	\resizebox{\linewidth}{!}{%
		\begin{tabular}{|l|c|c|c|c|c|c|c|c|c|c|c|c|c|c|c|c|}
			\hline
			& \multicolumn{4}{c|}{1-1} & \multicolumn{4}{c|}{1-N} & \multicolumn{4}{c|}{N-1} & \multicolumn{4}{c|}{N-N} \\
			\hline
			& MRR & H@1 & H@3 & H@10 & MRR & H@1 & H@3 & H@10 & MRR & H@1 & H@3 & H@10 & MRR & H@1 & H@3 & H@10 \\
			\hline
			TeRo   & .305 & .263 & .305 & .380 & .093 & .026 & .064 & \textbf{.263} & \textbf{.049} & \textbf{.018} & .038 & \textbf{.110} & .060 & .015 & .038 & .121 \\
			Ours  & \textbf{.314} & \textbf{.274} & \textbf{.317} & \textbf{.395} & \textbf{.093} & \textbf{.026} & \textbf{.071} & .250 & .048 & .016 & \textbf{.041} & .104 & \textbf{.069} & \textbf{.023} & \textbf{.043} & \textbf{.146} \\
			\hline
		\end{tabular}
	}
\end{table}

\begin{table}[htbp]
	\centering
	\caption{Performance on Wikidata12k by relation mapping properties. The best results are written bold.}
	\label{tab:type_wiki}
	\resizebox{\linewidth}{!}{%
		\begin{tabular}{|l|c|c|c|c|c|c|c|c|c|c|c|c|c|c|c|c|}
			\hline
			& \multicolumn{4}{c|}{1-1} & \multicolumn{4}{c|}{1-N} & \multicolumn{4}{c|}{N-1} & \multicolumn{4}{c|}{N-N} \\
			\hline
			& MRR & H@1 & H@3 & H@10 & MRR & H@1 & H@3 & H@10 & MRR & H@1 & H@3 & H@10 & MRR & H@1 & H@3 & H@10 \\
			\hline
			TeRo   & \textbf{.342} & \textbf{.249} & \textbf{.375} & \textbf{.522} & \textbf{.497} & \textbf{.426} & .520 & .634 & \textbf{.235} & \textbf{.136} & \textbf{.256} & \textbf{.438} & \textbf{.224} & .134 & \textbf{.245} & \textbf{.410} \\
			Ours  & .328 & .244 & .347 & .500 & .484 & .407 & \textbf{.521} & \textbf{.641} & .223 & .133 & .248 & .409 & .219 & \textbf{.139} & .237 & .370 \\
			\hline
		\end{tabular}
	}
\end{table}
Table 6, table 7, table 8 and table 9 show that the overall performance improvement of TeRoR on the ICEWS and YAGO11k datasets is consistently validated across most relation types. On ICEWS and YAGO11k, TeRoR maintains consistent positive gains across all four relation types: 1-1, 1-N, N-1, and N-N. Taking ICEWS05-15 as an example, the MRR for 1-1 relations increases from .475 to .491, Hits@10 for 1-N relations rises from .799 to .803, and the MRR improvement for N-1 relations reaches as high as 0.037. On Wikidata12k, TeRoR maintains comparable overall performance with TeRo. The comparative results verify that TeRoR is highly suitable for processing complex relational patterns, which effectively remedies the weakness of TeRo in multi-relational modeling.

\section{Conclusion}

We propose TeRoR to address the limitations of the temporal knowledge graph embedding model TeRo in handling complex relation mapping properties (1-N, N-1, N-N) and modeling temporal information. The core improvements of TeRoR lie in two aspects. First, the temporal evolution of entity embeddings is decomposed into independent complex-plane phase rotations for head and tail entities, which enhances the representation and utilization of temporal information. Second, relation-aware radius-constrained circular domains are introduced to define effective geometric boundaries for rotated and translated head entities, thereby precisely characterizing the mapping characteristics of various complex relations.

Experimental results on four standard datasets demonstrate that TeRoR outperforms existing baseline models and the original TeRo on most evaluation metrics, exhibiting strong capability for temporal link prediction. Ablation experiments further verify that both the independent entity rotation module and the radius-constrained circular domain module contribute significantly to model performance, and their collaboration can maximize the advantages of complex space rotation embedding. Analysis at the relation type level shows that TeRoR achieves consistent performance gains across 1-1, 1-N, N-1, and N-N relation types, fully validating the effectiveness of the radius constraint and phase-enhanced rotation mechanisms in modeling complex relations.

%
%
%
%

\bibliographystyle{splncs04}
\bibliography{ckwx}

\begin{thebibliography}{10}
\providecommand{\url}[1]{\texttt{#1}}
\providecommand{\urlprefix}{URL }
\providecommand{\doi}[1]{https://doi.org/#1}

\bibitem{auer2007dbpedia}
Auer, S., Bizer, C., Kobilarov, G., Lehmann, J., Cyganiak, R., Ives, Z.G.:
  Dbpedia: {A} nucleus for a web of open data. In: The Semantic Web, 6th
  International Semantic Web Conference, 2nd Asian Semantic Web Conference. pp.
  722--735. Springer (2007)

\bibitem{bollacker2008freebase}
Bollacker, K., Evans, C., Paritosh, P.K., Sturge, T., Taylor, J.: Freebase: a
  collaboratively created graph database for structuring human knowledge. In:
  Proceedings of the {ACM} {SIGMOD} International Conference on Management of
  Data. pp. 1247--1250. {ACM} (2008)

\bibitem{bordes2013translating}
Bordes, A., Usunier, N., Garc{\'{\i}}a{-}Dur{\'{a}}n, A., Weston, J.,
  Yakhnenko, O.: Translating embeddings for modeling multi-relational data. In:
  Advances in Neural Information Processing Systems 26: 27th Annual Conference
  on Neural Information Processing Systems 2013. pp. 2787--2795 (2013)

\bibitem{carlson2010toward}
Carlson, A., Betteridge, J., Kisiel, B., Settles, B., Jr., E.R.H., Mitchell,
  T.M.: Toward an architecture for never-ending language learning. In:
  Proceedings of the Twenty-Fourth {AAAI} Conference on Artificial
  Intelligence. pp. 1306--1313. {AAAI} Press (2010)

\bibitem{dasgupta2018hyte}
Dasgupta, S.S., Ray, S.N., Talukdar, P.P.: Hyte: Hyperplane-based temporally
  aware knowledge graph embedding. In: Proceedings of the 2018 Conference on
  Empirical Methods in Natural Language Processing. pp. 2001--2011. Association
  for Computational Linguistics (2018)

\bibitem{fabian2007yago}
Fabian, M., Gjergji, K., Gerhard, W., et~al.: Yago: A core of semantic
  knowledge unifying wordnet and wikipedia. In: 16th International world wide
  web conference. pp. 697--706 (2007)

\bibitem{garcia2018learning}
Garc{\'{\i}}a{-}Dur{\'{a}}n, A., Dumancic, S., Niepert, M.: Learning sequence
  encoders for temporal knowledge graph completion. In: Proceedings of the 2018
  Conference on Empirical Methods in Natural Language Processing. pp.
  4816--4821. Association for Computational Linguistics (2018)

\bibitem{goel2020diachronic}
Goel, R., Kazemi, S.M., Brubaker, M.A., Poupart, P.: Diachronic embedding for
  temporal knowledge graph completion. In: The Thirty-Fourth {AAAI} Conference
  on Artificial Intelligence, The Thirty-Second Innovative Applications of
  Artificial Intelligence Conference, The Tenth {AAAI} Symposium on Educational
  Advances in Artificial Intelligence. pp. 3988--3995. {AAAI} Press (2020)

\bibitem{ji2015knowledge}
Ji, G., He, S., Xu, L., Liu, K., Zhao, J.: Knowledge graph embedding via
  dynamic mapping matrix. In: Proceedings of the 53rd Annual Meeting of the
  Association for Computational Linguistics and the 7th International Joint
  Conference on Natural Language Processing of the Asian Federation of Natural
  Language Processing. pp. 687--696. The Association for Computer Linguistics
  (2015)

\bibitem{lacroix2018canonical}
Lacroix, T., Usunier, N., Obozinski, G.: Canonical tensor decomposition for
  knowledge base completion. In: Proceedings of the 35th International
  Conference on Machine Learning. pp. 2869--2878. {PMLR} (2018)

\bibitem{lautenschlager2015icews}
Lautenschlager, J., Shellman, S., Ward, M.: Icews event aggregations  (2015)

\bibitem{leblay2018deriving}
Leblay, J., Chekol, M.W.: Deriving validity time in knowledge graph. In:
  Companion of the The Web Conference 2018 on The Web Conference 2018. pp.
  1771--1776. {ACM} (2018)

\bibitem{leetaru2013gdelt}
Leetaru, K., Schrodt, P.A.: Gdelt: Global data on events, location, and tone,
  1979--2012. In: ISA annual convention. vol.~2, pp. 1--49. Citeseer (2013)

\bibitem{lin2015learning}
Lin, Y., Liu, Z., Sun, M., Liu, Y., Zhu, X.: Learning entity and relation
  embeddings for knowledge graph completion. In: Proceedings of the
  Twenty-Ninth {AAAI} Conference on Artificial Intelligence. pp. 2181--2187.
  {AAAI} Press (2015)

\bibitem{mahdisoltani2014yago3}
Mahdisoltani, F., Biega, J., Suchanek, F.M.: {YAGO3:} {A} knowledge base from
  multilingual wikipedias. In: Seventh Biennial Conference on Innovative Data
  Systems Research. www.cidrdb.org (2015)

\bibitem{nickel2016holographic}
Nickel, M., Rosasco, L., Poggio, T.A.: Holographic embeddings of knowledge
  graphs. In: Proceedings of the Thirtieth {AAAI} Conference on Artificial
  Intelligence. pp. 1955--1961. {AAAI} Press (2016)

\bibitem{nickel2011three}
Nickel, M., Tresp, V., Kriegel, H.: A three-way model for collective learning
  on multi-relational data. In: Proceedings of the 28th International
  Conference on Machine Learning. pp. 809--816. Omnipress (2011)

\bibitem{sun2019rotate}
Sun, Z., Deng, Z., Nie, J., Tang, J.: Rotate: Knowledge graph embedding by
  relational rotation in complex space. In: 7th International Conference on
  Learning Representations. OpenReview.net (2019)

\bibitem{trouillon2016complex}
Trouillon, T., Welbl, J., Riedel, S., Gaussier, {\'{E}}., Bouchard, G.: Complex
  embeddings for simple link prediction. In: Proceedings of the 33nd
  International Conference on Machine Learning. pp. 2071--2080. {JMLR} Workshop
  and Conference Proceedings, JMLR.org (2016)

\bibitem{wang2014knowledge}
Wang, Z., Zhang, J., Feng, J., Chen, Z.: Knowledge graph embedding by
  translating on hyperplanes. In: Proceedings of the Twenty-Eighth {AAAI}
  Conference on Artificial Intelligence. pp. 1112--1119. {AAAI} Press (2014)

\bibitem{xu2020tero}
Xu, C., Nayyeri, M., Alkhoury, F., Yazdi, H.S., Lehmann, J.: Tero: {A}
  time-aware knowledge graph embedding via temporal rotation. In: Proceedings
  of the 28th International Conference on Computational Linguistics. pp.
  1583--1593. International Committee on Computational Linguistics (2020)

\bibitem{xu2020temporal}
Xu, C., Nayyeri, M., Alkhoury, F., Yazdi, H.S., Lehmann, J.: Temporal knowledge
  graph completion based on time series gaussian embedding. In: The Semantic
  Web - {ISWC} 2020 - 19th International Semantic Web Conference. pp. 654--671.
  Springer (2020)

\bibitem{yang2014embedding}
Yang, B., Yih, W., He, X., Gao, J., Deng, L.: Embedding entities and relations
  for learning and inference in knowledge bases. In: 3rd International
  Conference on Learning Representations (2015)

\bibitem{zhang2019quaternion}
Zhang, S., Tay, Y., Yao, L., Liu, Q.: Quaternion knowledge graph embeddings.
  In: Advances in Neural Information Processing Systems 32: Annual Conference
  on Neural Information Processing Systems 2019. pp. 2731--2741 (2019)

\end{thebibliography}

\end{document}